\documentclass[letterpaper, 10 pt, journal, twoside]{IEEEtran}  

\IEEEoverridecommandlockouts                              

\usepackage{cite}
\usepackage{amsmath,amssymb,amsfonts}
\usepackage{algorithmic}
\usepackage{graphicx}
\usepackage{textcomp}
\usepackage{needspace}
\usepackage{tensor}
\usepackage{bm}
\usepackage{graphicx}
\usepackage{caption}
\usepackage{color}
\usepackage{midfloat}
\usepackage{float}
\usepackage{rotating}
\usepackage{multirow}
\usepackage{hhline}
\usepackage{booktabs}
\usepackage{makecell}
\usepackage{longtable}
\usepackage{colortbl}
\usepackage{array}
\usepackage{paralist}
\usepackage{multicol}
\usepackage{multirow}
\usepackage{chngcntr}
\usepackage{inconsolata}
\usepackage{arydshln,leftidx,mathtools}
\usepackage{textcomp}
\usepackage{microtype}
\usepackage{stfloats}
\usepackage{microtype}
\usepackage{lipsum}
\usepackage{needspace}

\usepackage[compact]{titlesec}
\titlespacing{\section}{0pt}{1.5ex}{1ex}
\titlespacing{\subsection}{0pt}{1ex}{0ex}
\titlespacing{\subsubsection}{0pt}{0.5ex}{0ex}

\title{\LARGE \bf The Role of Functional Muscle Networks in Improving Hand Gesture Perception for Human-Machine Interfaces}

\author{Costanza Armanini$^{*1}$, 
        Tuka Alhanai$^{2}$, 
	Farah E. Shamout$^{2,3,4}$,
	S. Farokh Atashzar$^{3,5,6}$
\thanks{The material presented in this paper is supported in part by the Center for AI and Robotics (CAIR) at New York University Abu Dhabi: Award CG010. The work is also supported in part by the US National Science Foundation under Grant No. 2229697 and No. 2121391. The work is also supported by an award from Mathworks.

Corresponding author {Costanza Armanini \tt\footnotesize ca3072@nyu.edu}.}
\thanks{$^{1}$ Center for Artificial Intelligence and Robotics (CAIR), New York University Abu Dhabi (NYUAD), Abu Dhabi, UAE}%
\thanks{$^{2}$ Division of Engineering, New York University Abu Dhabi (NYUAD), Abu Dhabi, UAE}
\thanks{$^{3}$ Department of Biomedical Engineering, New York University (NYU), New York, USA}
\thanks{$^{4}$ Department of Computer Science and Engineering, New York University (NYU), New York, USA}
\thanks{$^{5}$ Department of Electrical and Computer Engineering, New York University (NYU), New York, USA}
\thanks{$^{6}$ Department of Mechanical and Aerospace Engineering, New York University (NYU), New York, USA}

\title{\LARGE \bf The Role of Functional Muscle Networks in Improving Hand Gesture Perception for Human-Machine Interfaces}
}

\begin{document}

\maketitle

\begin{abstract} 
Developing accurate hand gesture perception models is critical for various robotic applications, particularly for interactive robots and neurorobots. Such models enable more intuitive and effective communication between humans and machines, directly impacting neurorobotics. Recently, surface electromyography (sEMG) has been widely explored due to its rich informational context and accessibility when combined with advanced machine learning approaches and wearable systems. The literature has explored numerous approaches to boost performance while ensuring robustness for neurorobots using sEMG. However, these efforts often result in models requiring high processing power, large datasets, and less scalable solutions. This paper addresses this challenge by proposing the decoding of muscle synchronization rather than individual muscle activation. We investigate coherence-based functional muscle networks as the core of the computational process for our perception model. We hypothesize that functional synchronization between muscles and the corresponding graph-based network of muscle connectivity encodes contextual information about intended hand gestures. This can be decoded using shallow machine learning approaches (e.g., support vector machines) without the need for deep temporal networks, which demand high computational resources. Our technique could impact myoelectric control of neurorobots by significantly reducing computational burdens and enhancing efficiency. The proposed approach is benchmarked using the Ninapro DB-2 database, which contains 12 EMG signals from 40 subjects performing 17 hand gestures. The approach achieves an overall accuracy of $\mathbf{85.1\%}$, demonstrating improved performance compared to existing methods while requiring much less computational power. The results support the hypothesis that a coherence-based functional muscle network encodes critical information related to gesture execution, significantly enhancing hand gesture perception with potential applications for neurorobotic systems and interactive machines.

\end{abstract}


\section{INTRODUCTION}
\label{sec:introduction}

Neurorobotics and interactive robotics stand out among the research fields in which Hand Gesture Recognition (HGR) models have been predominantly applied. Such models are at the core of the control strategies, as they facilitate intuitive and effective communication between humans and machines, which is crucial for the seamless interaction. Surface electromyography (sEMG) has been widely studied for their applications in control of prosthetic devices, neurorobotics, and human-computer interfaces \cite{Sayin_2018},  \cite{Wei_2019}, \cite{Kumar_2022}, \cite{Raez_2006}. This is driven by the potential in recognizing and predicting hand gestures in a wearable, non-invasive way. 

When combined with Machine Learning (ML), sEMG biosignals offer great potential for developing accurate perception models, enhancing their interpretation and enabling more effective hand gesture recognition models \cite{JaramilloYanez_2020}, \cite{Kim_2008}. In these approaches, the features are usually extracted from the time (such as root mean square, variance, mean absolute value, zero crossings, histogram) or frequency (such as short-time Fourier transform, cepstral coefficients) domain of the sEMG, and these can then be fed into conventional classifiers, such as Support Vector Machines (SVMs) or Linear Discriminant Analysis (LDA) \cite{Shehata_2021}, \cite{Phinyomark_2013}, \cite{Tavakoli_2018}, \cite{Gijsberts_2014}. One of the main limitations of conventional ML methods is in the simplicity of the extracted features and the utilized models. Thus, they can struggle to maintain high accuracy as the complexity and variety of gestures increase. With the advent of big data, researchers have leaned towards Deep Learning (DL) approaches, which have shown promise in improving the performance of sEMG-based gesture recognition tasks. These models leverage vast amounts of data and sophisticated neural network architectures to achieve superior accuracy and robustness. In particular, Convolutional Neural Networks (CNN) remove the need for feature engineering, improving the model performance \cite{Wei_2019}, \cite{Ding_2018}. Long Short-Term Memory (LSTM) Recurrent Neural Networks (RNN) have also been considered for their potential in capturing the gesture temporal dynamics \cite{Cifuentes_2019}, \cite{Jabbari_2020}.
More recently, our team has shown the potential of more advanced DL models, such as vision transformers, combined with transfer learning, which enhanced the performance and generalizability of such neural interfaces \cite{Erin_JSTSP}, \cite{Erin_INSTRUM}. Despite their potential, DL models come with their own set of challenges. Most of them were originally designed for text processing applications and are computationally heavy, requiring extensive datasets and substantial processing power, which can be a barrier to their practical implementation in real-time systems. 
In this letter, we introduce a novel approach centered on the concept of functional muscle networks, with the aim of providing an accurate HGR model that is capable of balancing performance and computational efficiency.  

\begin{figure*}[b!]
     \centering
    \includegraphics[width=0.9\linewidth]{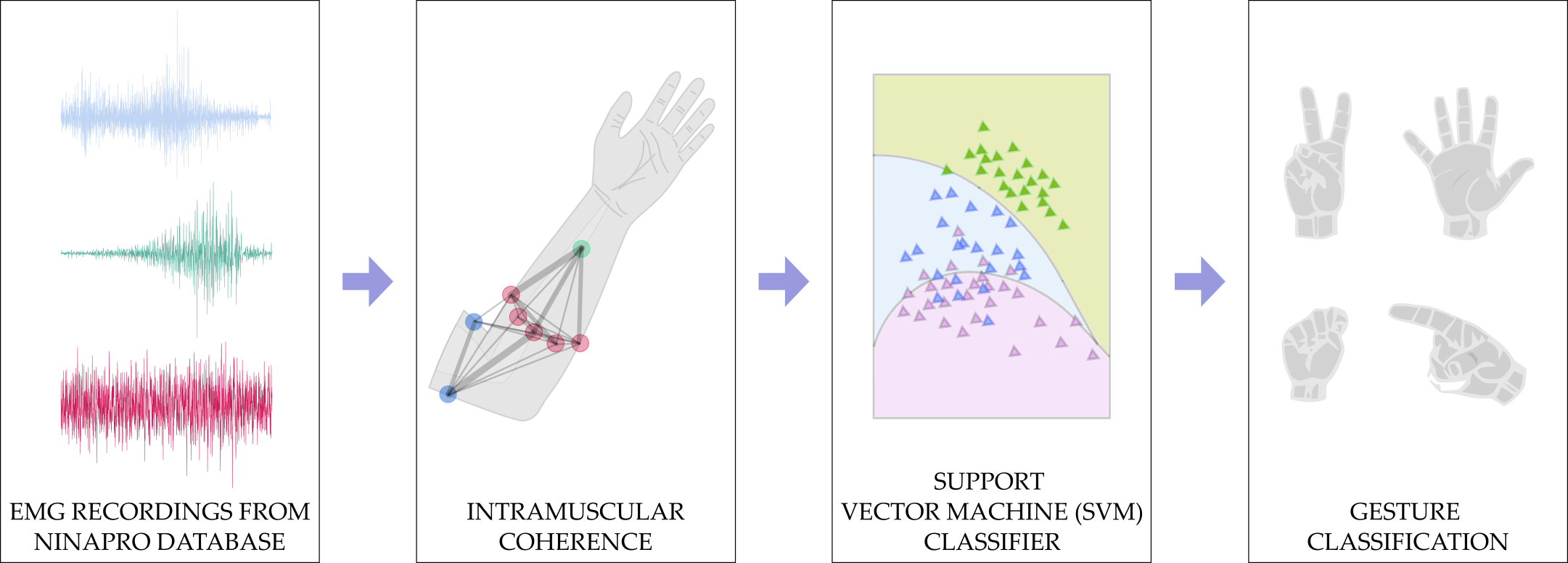}
    \caption{A schematic of the presented research. The NinaPro DB2 (Exercise B) dataset provides sEMG recordings of $17$ movements performed by $40$ healthy subjects. After minor pre-processing, the recordings are used to obtain muscular coherence values, which are then used as features for a shallow support vector machine gesture classifier.}
    \label{fig::WorkFlow}
 \end{figure*}
 
Recently, the concept of coherence-based functional muscle networks from neuroscience has aimed to model the synergistic coupling between various muscles, mainly for application in diagnosis and biomarkers of neuromusculoskeletal disorders, such as the pathophysiology of Parkinson's disease \cite{Sarnthein_2007}, the alterations in functional muscular connectivity for stroke survivors \cite{Houston_2021}, new rehabilitation protocols \cite{Tung_2013}, \cite{OKeeffe_SciRep_2022}, or the effect of fatigue during repetitive exercise \cite{Garcia_Retortillo_2023}, \cite{OKeeffe_JBHI}. Muscular coherence has also been investigated to study the synchronization and control techniques of the human body while performing simple daily life chores, such as postural tasks \cite{Boonstra_2015}, walking \cite{Kerkman_Frontiers2020}, or to assess vocal performance and voice disorders \cite{Rory_TBME2022}. 
Inspired by concepts from graph theory and brain connectivity analysis, functional muscle networks represent an emerging tool for decoding how the neural drive coming from the the Central Nervous System (CNS) is distributed synergistically between various muscles to control the muscular skeletal system. These networks can provide important insights into the CNS control and orchestration of muscle activation while performing functional tasks. 
It is widely accepted that when the human body performs a movement, the CNS  encodes the function into controlling a sequence of \textit{collaborative} muscle functions (such as selecting from a library of actions) \cite{Bizzi_2013}. This concept is classically studied in the literature as synergistic control of muscles, which has been recently extended to the concept of synergistic functional networks that would give a holistic view of the distribution of neural drive across different muscles while doing different functional tasks, \cite{DAvella_2013}. These concepts highlight that the nervous and musculoskeletal systems work together to coordinate the muscle synergies necessary for executing complex movements and behaviors.  Muscular coherence represents the degree of synchronization between muscles acting in harmony at various frequencies to execute different functional tasks. It should be highlighted that the term \textit{coherence} refers to a statistical measure used to examine the relationship between two signals at various frequencies and \textit{muscular coherence} specifically studies the synchronization between different \textit{muscle} signals \cite{Laine_2017}. Here, the connectivity metric is the Magnitude Squared Coherence (MSC), which indicates the linear association between two signals in the frequency domain. The MSC value ranges from $0$ to $1$, where $MSC = 1$ indicates perfect linear correlation (the signals are identical) at that particular frequency, while $MSC = 0$, indicates that the signals are completely uncorrelated at that frequency. 

In this paper, we present a machine learning classifier based on the analysis of the conference of functional muscles representing the orchestration of functional muscles when performing a hand gesture. The visual workflow of the proposed approach is given in Fig. \ref{fig::WorkFlow}. To investigate our hypothesis we benchmark our model using a publicly available dataset - NinaPro \cite{Ninapro_Original}. In this regard, we examine the coherence based functional muscular network patterns for $40$ able-bodied subjects while performing $17$ distinct hand movements.  Frequency-based features from functional muscular networks are processed by a shallow SVM for gesture classification. 
The results supported the central hypothesis of this paper as it demonstrates that coherence based functional muscular networks encode critical information related to gesture execution, allowing for the development of a low-dimensional gesture classifier that effectively leverages the harmony of function across various muscles rather than the magnitude of action at individual nodes. This approach also improves our understanding of muscular coordination and showcases the practical applications of muscular coherence networks in creating efficient and accurate gesture perception models. This method could not only streamline the classification process but also enhance its efficiency, making it a viable option for real-time applications, including neurorobots and interactive systems. Moreover, the results provide deeper insights into the underlying neuromuscular processes and enable more advanced and interpretable models.

The remaining parts of the paper are organized as follows. In Section \ref{sec::Data} we present the data and their pre-processing, while Section \ref{sec::Coherence} details the coherence analysis and the study of muscular networks. Section \ref{sec::Model} uses these results to develop the support vector machine gesture classifier and presents the results and their comparison with the existing literature. Finally, in Section \ref{sec::Concl}, we draw conclusions and discuss the future prospects of the presented work.
 
\section{Data and Preprocessing}
\label{sec::Data}

Different sEMG datasets have been presented and made publicly available in the literature \cite{Phinyomark_2018} \cite{Kaczmarek_2019} \cite{Ozdemir_2022} \cite{Nunez}, and among these, the NinaPro initiative \cite{Ninapro_Original} stands out for its popularity in the scientific and robotics community.  Short for Non-Invasive Adaptive Prosthetics, NinaPro is an extensive research project that provides a comprehensive collection of sEMG recordings gathered from both intact and trans-radial hand amputees performing a range of hand and wrist gestures. It is composed of $10$ collections, varying in the number of subjects, movements, and acquiring methods. 
\begin{figure*}[t!]
    \centering
    \includegraphics[width=0.9\linewidth]{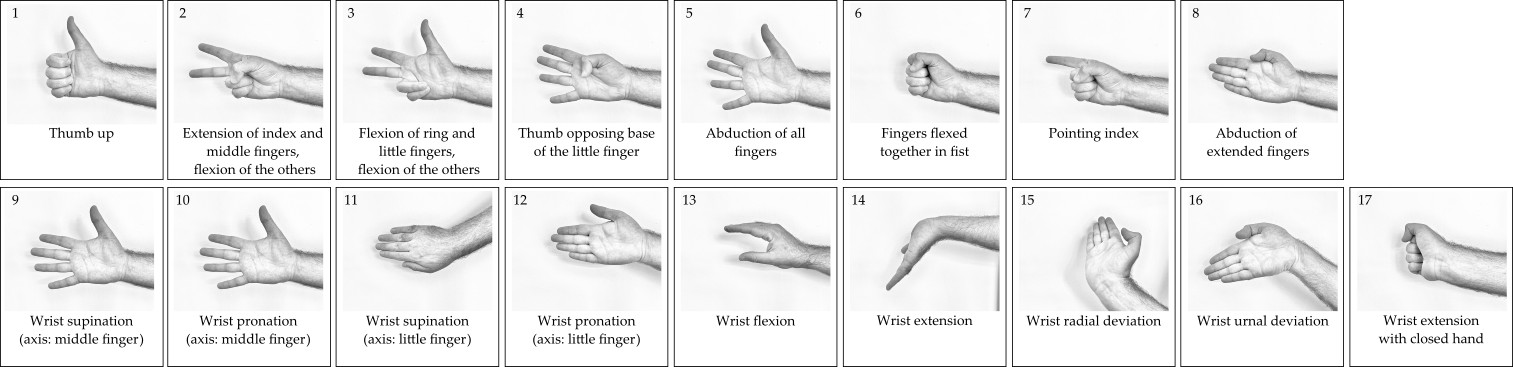}
    \caption{The $17$ gestures performed in the Ninapro dataset DB2 (exercise set B).}
    \label{fig::Gestures}
 \end{figure*}

In particular, we employ the Ninapro DB-2 dataset, which contains sEMG, inertial, kinematic, and force data from $40$ able-bodied subjects ($28$ males, $12$ females; age $29.9 \pm 3.9$ years; $34$ right-handed and $6$ left-handed). Specifically, the kinematic data were gathered with a Cyberglove 2 data glove, while the sEMG data were collected using $12$ Delsys Trigno electrodes, placed as follows: eight electrodes were evenly spaced around the forearm near the radio humeral joint, two electrodes were placed on the primary activity sites of the flexor and extensor digitorum muscles, and two electrodes were placed on the main activity spots of the biceps and triceps muscles. The sEMG signals were sampled at a $2000$ Hz frequency. 

During data collection, the subjects were asked to follow several movements represented by movies shown on a laptop screen. All subjects performed the movements with their right hand for $5$ seconds, followed by $3$ seconds of rest, and the protocol included $6$ repetitions.  While the full dataset includes $49$ hand and wrist movements, here we only focus on the $17$ gestures from Exercise B, which are presented in Fig. \ref{fig::Gestures}.

\begin{figure*}[t!]
     \centering
    \includegraphics[width=\linewidth]{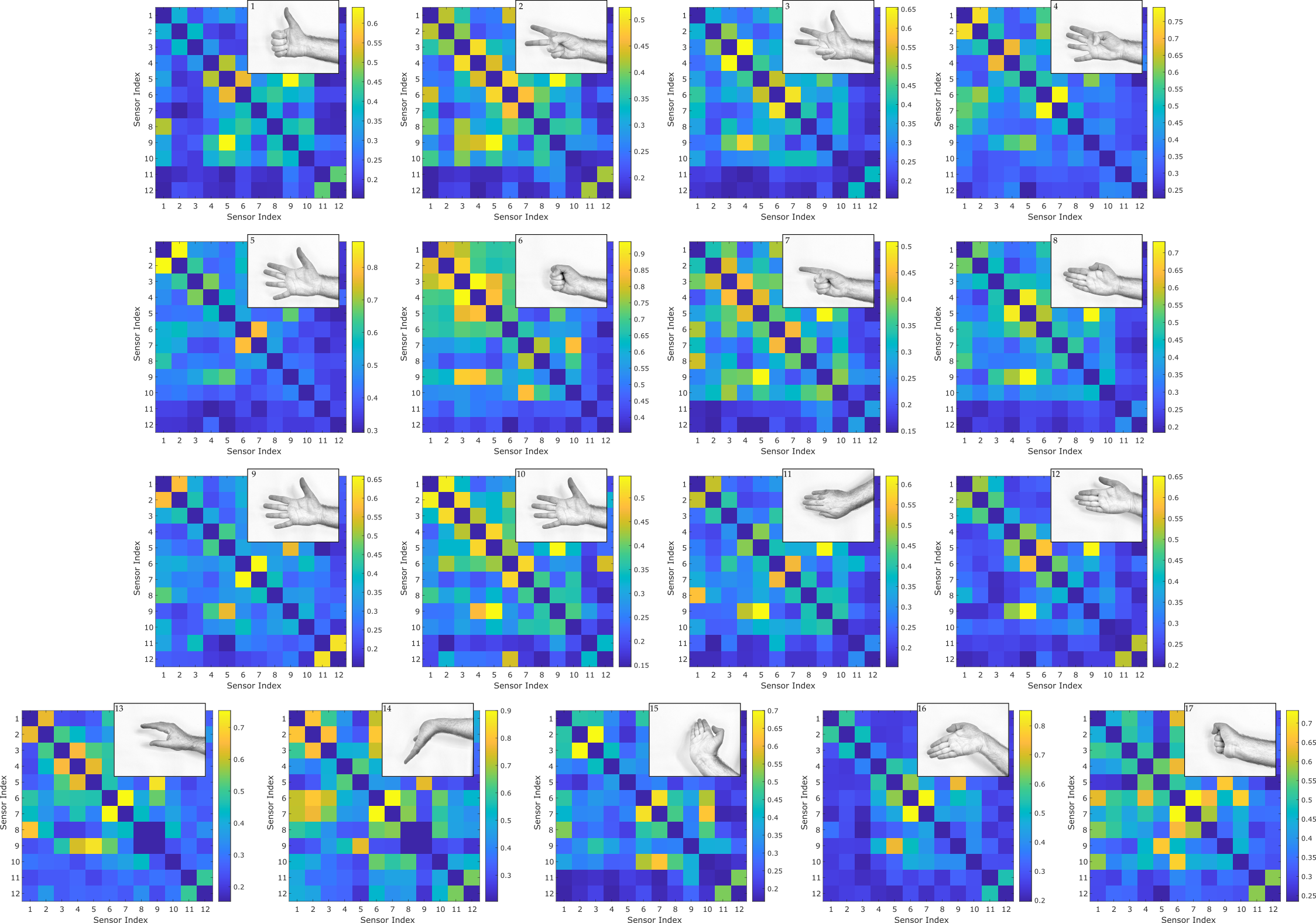}
    \caption{Coherence matrices obtained for each gesture from Subject 19. Each matrix is $12 \times 12 = 144$ squares, representing the corresponding signal combination's median coherence values across all repetitions. The color is proportional to the pairwise coherence. Since interconnectivity is uni-directed, the matrices are all symmetric. The diagonal terms, corresponding to self-pairing signals,  were manually set to $0$ in order to avoid overshadowing other connectivity patterns with low MSC values.}
    \label{fig::MSC}
 \end{figure*}
 
\subsection{Pre-processing}
We use each gesture's full data duration of $5$ seconds, including the transient phase (which is represented by the first $1$ second for the movement activation \cite{Eion_RAL2022}) and the steady-state of the contraction, which corresponds to the remaining $4$ seconds. The train and test sets for gesture classification are split based on the repetition number, as in \cite{Ninapro_Original} and other existing studies that we use as baseline: repetitions $1$, $3$, $4$, and $6$ are used as the training set, while repetitions $2$ and $5$ are used as the test set.

To ensure the quality and usability of the sEMG data, a series of pre-processing steps are performed using digital filters. The pre-processing involves two primary filtering operations: 
\begin{itemize}
    \item \textbf{Bandpass Filtering}: a fourth-order Butterworth filter is designed with the specified passband ($10$ Hz to $900$ Hz) to retain the relevant sEMG signal frequencies while attenuating those outside this range. 
    \item \textbf{Notch Filtering}: a notch filter removes the powerline interference at $50$ Hz.
\end{itemize}

Zero-phase filtering is also applied to avoid phase distortion. We then applied Z-score normalization, where the data is normalized to have a mean of $0$ and a standard deviation of $1$, based on the training set only. This ensures that the data is on a common scale, which is essential for effective analysis and comparison. 

\section{Muscolar Coherence Analaysis}
\label{sec::Coherence}
Muscular coherence indicates how well one muscle signal's frequency components align with another's over time. High coherence values suggest that the muscles are working together in a coordinated manner, while low coherence values imply lower synergy. The coherence between signal pairs can be quantified using the Magnitude-Squared Coherence (MSC). Given two signals $x(t)$ and $y(t)$, their MSC is defined as:
\begin{equation}
\label{eq:MSC}
    MSC(f) = \frac{|{P}_{xy}(f)|^2}{P_{xx}(f) \cdot P_{yy}(f)} \; ,
\end{equation}

where:
\begin{itemize}
    \item $P_{xy}(f)$ is the cross-spectral density of signals $x$ and $y$ at frequency $f$;
    \item $P_{xx}(f)$ and $P_{yy}(f)$ are the power-spectral densities of signals $x$ and $y$ at frequency $f$.
\end{itemize}

The MSC value ranges from $0$ to $1$, where $MSC = 1$ indicates perfect linear correlation at that particular frequency (or, in other words, the signals are identical at this frequency), while $MSC = 0$, indicates no linear correlation at a particular frequency (the signals are completely uncorrelated at this frequency). By analyzing MSC values across different muscle pairs and frequencies, we can gain insights into the coordination patterns of muscle activity during the considered gestures.
\begin{figure*}[t!]
     \centering
    \includegraphics[width=\linewidth]{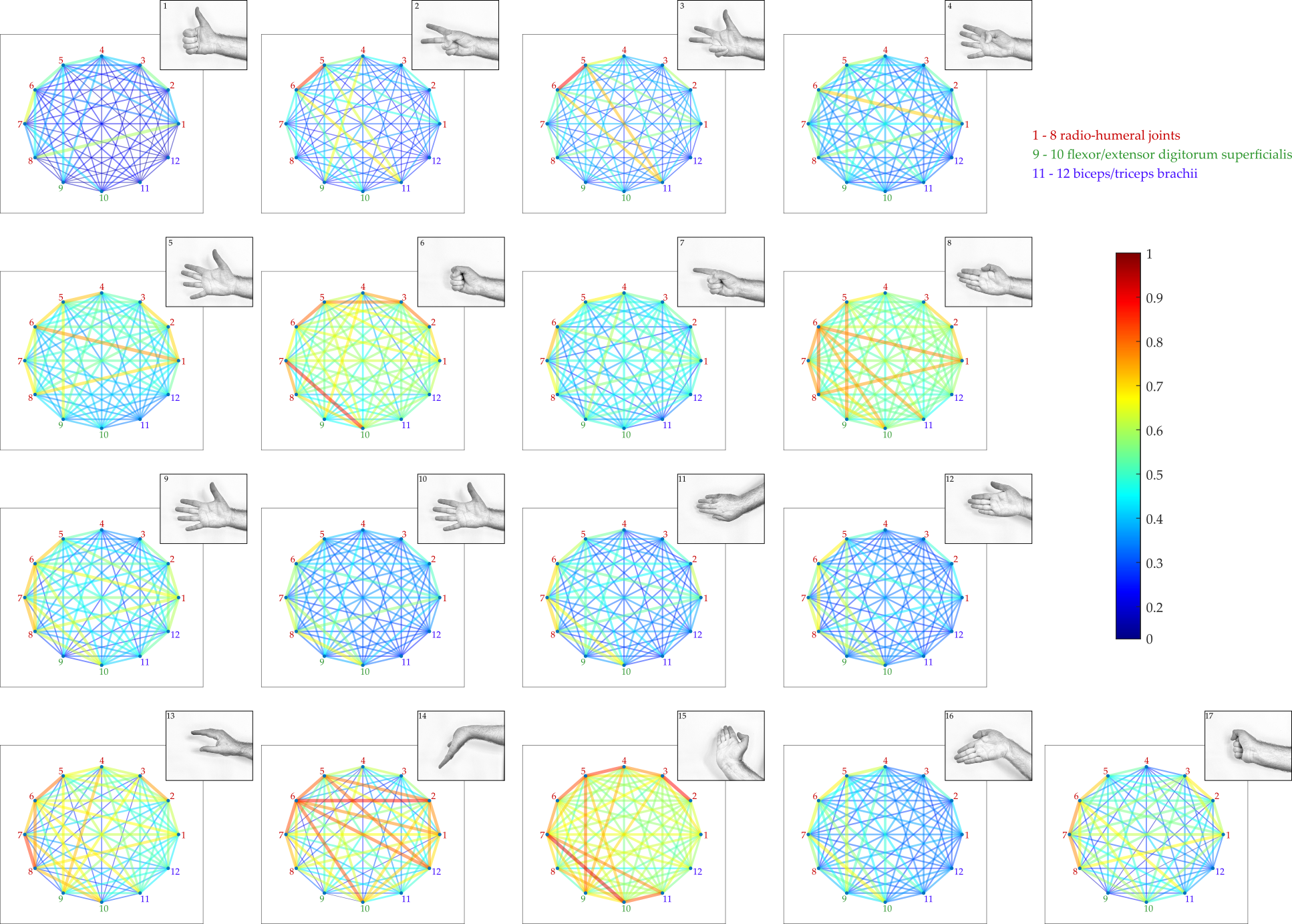}
    \caption{Coherence muscle networks from Subject 1 obtained for each hand gesture. Each node in the network represents a sensor, while the color of the edges between the nodes denotes the strength of the coherence between them, where blue corresponds to $MSC = 0$ (no coherence) and red corresponds to $MSC = 1$ (maximum coherence). }
    \label{fig::Networks}
 \end{figure*}

To compute the power spectral densities in equation \ref{eq:MSC}, the Welch's method is applied with a window size of $600$ samples (which, given the sampling frequency of $2$ kHz, corresponds to $300$ ms timesteps) and $50\%$ overlap. Finally, we obtained the adjacent coherence matrices containing the MSC values over the frequency domain for each signal combination. 

Fig.  \ref{fig::MSC} presents the results obtained for Subject $19$ for all $17$ gestures. Each matrix contains $12 \times 12 = 144$ squares, representing the corresponding signal combination's median coherence values across all repetitions. The color is proportional to the pairwise coherence, where blue and yellow colors correspond to $MSC = 0$ (no coherence) and $MSC = 1$ (maximum coherence), respectively. It should be noted that since interconnectivity is uni-directed, the matrices are all symmetric. Moreover, the diagonal terms correspond to self-pairing signals, i.e., $x=y$ in equation \ref{eq:MSC}, which necessarily leads to $MSC=1$ (the signals are identical). These values were manually set to $0$ in order to avoid overshadowing other connectivity patterns with low MSC values. 

Fig. \ref{fig::Networks} represents the coherence networks obtained from Subject 1. Each node in the network represents a sensor, while the edges between the nodes represent the coherence values between them, and the color intensity of the edges encodes the strength of the coherence.

From both figures, it can be noted that the wrist extension gesture (number $14$) presents the highest values of muscular coherence, followed by gestures $5$ (abduction of all fingers), $6$ (fingers together in a fist), $8$ (abduction of extended fingers), $13$ (wrist flexion) and $15$ (wrist radial deviation) with different levels of MSC depending on the Subject. 

\section{Gesture Classification Model} 
\label{sec::Model}

In this section, we introduce the proposed model, discussing previous works that used the same dataset and that represent our baseline. The NinaPro DB-2 dataset has been widely employed for sEMG-based gesture classification through both traditional machine learning methods, such as random forest, SVM, k-nearest neighbors \cite{Ninapro_Original}, and advanced deep learning techniques, including CNNs \cite{Atzori_2016}, \cite{Ding_2018}, LSTMs \cite{Sun_2021}, hybrid approaches \cite{Gulati_2021}, and the dilated efficient Capsule Neural Network (CapsNet) \cite{Eion_RAL2022}. Table \ref{table:Comparison} outlines the characteristics and accuracy of the approaches using the same dataset that we consider here (Exercise B with $17$ gestures), establishing the baseline for our method. In this paper, we aim to demonstrate that, leveraging the concept of muscular coherence can achieve comparable or superior classification results. Our approach focuses on extracting meaningful features from the magnitude-squared coherence of the sEMG signals, reducing the complexity of the model without compromising its accuracy. This can be accomplished using a traditional machine learning model, such as an SVM.  

After obtaining the MSC matrices in Section \ref{sec::Coherence}, the computational cost of the algorithm is further reduced by considering the mean of the MSC values over the frequency domain. This process results in a $102 \times 132$ feature matrix. The number of rows ($102$) is determined by the product of $17$ movements and $6$ repetitions. For example, rows $1-6$ contain the information from the $6$ repetitions of gesture $1$, rows $7-12$ correspond to gesture $2$, and so on. Each row contains the mean coherence matrices for a specific movement and repetition combination, arranged in a vector format without the diagonal terms where $MSC = 1$. This results in $132$ columns: $12 \times 12 = 144$ possible signal combinations, minus the $12$ diagonal terms.
\begin{table*}[t!]
\centering
 \caption{Average accuracy, precision, recall, F1 score and area under the ROC curve across all subjects for each considered movement.}
\begin{tabular}{ |c|l|c|c|c|c|c| } 
 \hline
\multicolumn{2}{|c|}{\textbf{Gesture}} & \textbf{Accuracy} [$\textbf{\%}$] & \textbf{Precision} [$\textbf{\%}$] & \textbf{Recall} [$\textbf{\%}$] & \textbf{F1 Score} [$\textbf{\%}$] & \textbf{AUC}  [$\mathbf{10^{-2}}$] \\
 \hline
1 & Thumbs up & 91.3 & 87.3 & 91.3 & 87.6 & 97.8\\
 \hline
2 & Index and Middle Finers Exten., Flex & 83.6 & 85.4 & 83.8 & 81.9 & 95.7\\
 \hline
3 & Ring and Little Fingers Flex., Remaining  Exten. & 82.5 & 80.6 & 82.5 & 78.4 & 97.5\\
 \hline
4 & Thumb Opposing Base of Little Finger & 85.0 & 82.9 & 85.0 & 81.3 & 98.3\\
 \hline
5 & Abduction of All Fingers & 82.5 & 77.2 & 82.5 & 76.7 & 98.3\\
 \hline
6 & Fingers Flexed in Fist & 87.5 & 92.5 & 87.5 & 88.1 & 98.0\\
 \hline
7 & Pointing Index & 83.8 & 76.7 & 83.8 & 78.1 & 97.1\\
 \hline
8 & Abduction of Extended Fingers & 83.8 & 87.9 & 83.8 & 83.4 & 95.9\\
 \hline
9 & Wrist supination (axis Middle Finger) & 77.5 & 85.8 & 77.5 & 79.2 & 94.4 \\
 \hline
10 & Wrist pronation (axis Middle Finger) & 86.3 & 86.7 & 86.3 & 84.4 & 97.9\\
 \hline
11 & Wrist supination (axis Little Finger) & 87.5 & 85.8 & 87.5 & 84.3 & 96.6\\
 \hline
12 & Wrist pronation (axis Little Finger)  & 86.2 & 79.2 & 86.3 & 81.4 & 97.4\\
 \hline
13 & Wrist Flexion & 85.0 & 92.5 & 85.0 & 86.8 & 96.4 \\
 \hline
14 & Wrist Extension & 80.0 & 97.9 & 80.0 & 85.3 & 93.3\\
 \hline
15 & Wrist Radial Deviation & 76.2 & 84.6 & 76.3 & 78.6 & 91.7\\
 \hline
16 & Wrist Ulnar Deviation & 93.8 & 94.6 & 93.8 & 92.9 & 99.8\\
 \hline
17 & Wrist Extension with Closed Hand Deviation & 93.8 & 96.2 & 93.8 & 93.5 & 98.5 \\
 \hline
\multicolumn{2}{|c|}{\textbf{Average}}  & $\textbf{85.1}$ & $\textbf{86.7}$ & $\textbf{85.1}$ & $\textbf{83.6}$ & $\textbf{96.8}$\\
 \hline
\end{tabular}
 \label{table:Accuracy}
\end{table*}

The feature matrix is then employed to train a polynomial support vector machine model. To ensure a fair comparison with the literature, we follow the standard approach of using repetitions $1$, $3$, $4$, and $6$ for training, reserving the remaining two for testing. Given the limited dataset and the absence of a proper validation set, we employed a 3-fold cross-validation to perform a grid-search tuning of the hyperparameters of the SVM, i.e. the order of the polynomial kernel and the regularization parameter $\mathcal{C}$. This process led us to select a polynomial kernel of order $2$ and a $\mathcal{C}$ value of $10$, which provided the best performance in our cross-validation experiments.

\subsection{Results}
\begin{figure*}[b!]
     \centering
    \includegraphics[width=0.8\linewidth]{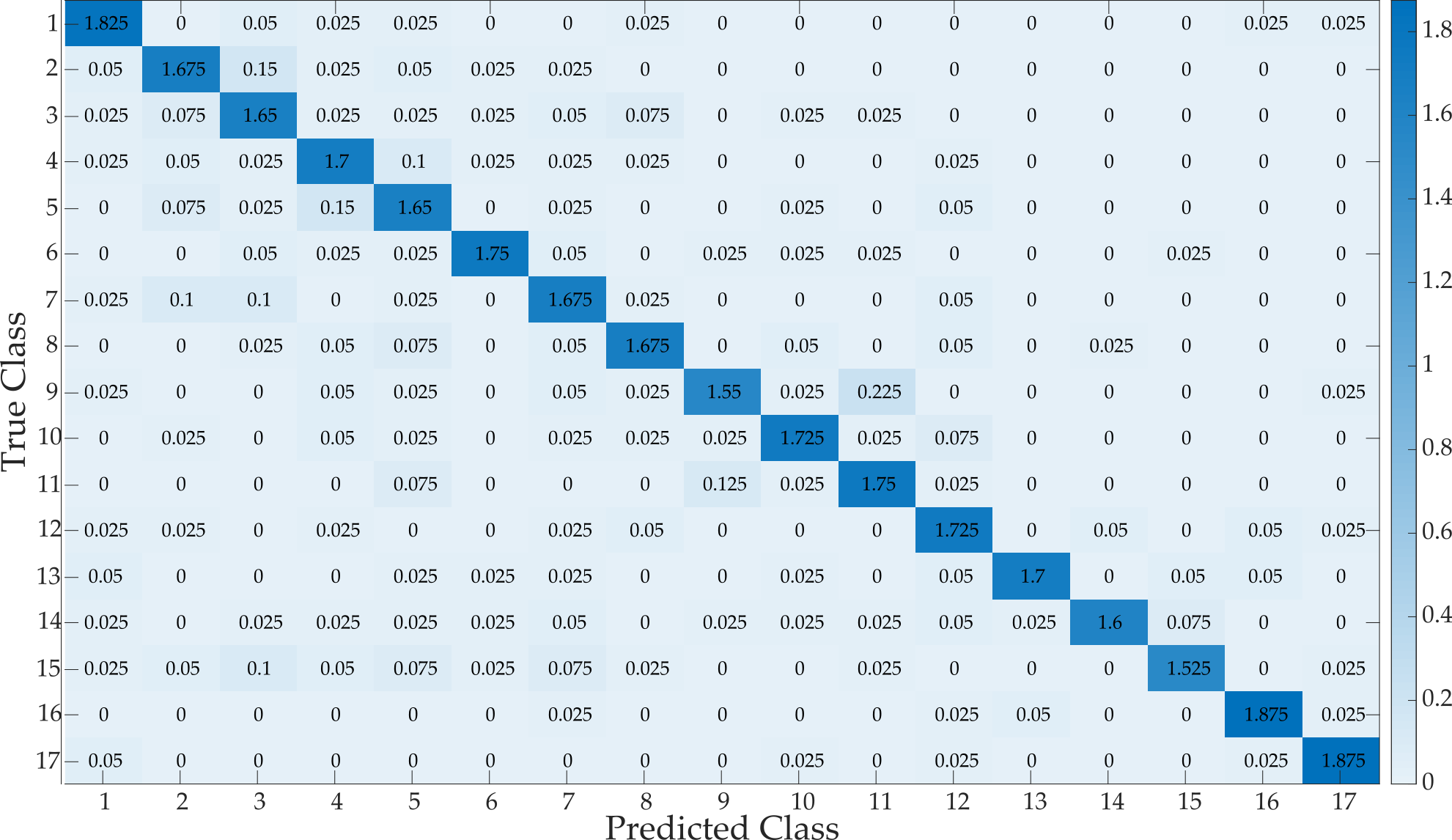} 
        \caption{Average confusion matrix across all subjects. On average, $28.9$ samples were correctly predicted out of the total $34$ samples ($2$ repetitions for $17$ gestures). It can be noted that classes presenting similar MSC values, such as Class 15 and Class 9, are the most likely confused.}
    \label{fig::ConfMatrix}
 \end{figure*}

The SVM classifier is trained and tested on each subject individually. Table \ref{table:Accuracy} presents the average accuracy, precision, recall, F1 score and the Area Under the receiver operating characteristic curve Curve (AUC) obtained for each considered gesture. Finally, Fig. \ref{fig::ConfMatrix} presents the average confusion matrix across all subjects. For each gesture, the test set consists of $2$ repetitions. On average, $28.9$ samples were correctly predicted out of the total $34$ samples. Class 9 (wrist supination along the middle finger axis) presents one of the highest misclassification rates, and it was mostly confused with Class 11 (wrist supination along the little finger axis). This could also be predicted from Figs. \ref{fig::MSC} and \ref{fig::Networks}, where these two classes present similar values of MSC. Class 15 (wrist radial deviation) was also confused with Class 3 (ring and little fingers flexion with the extension of the remaining ones), and, from the results of Subject 19 in Fig. \ref{fig::MSC}, it can be noted that the two corresponding MSC matrices are, in fact, similar. 

While using a shallow classifier with a low computational cost, the results are comparable, and sometimes superior, to the ones summarized in Table \ref{table:Comparison}, which includes the accuracy that was obtained with the same dataset, with state-of-the-art techniques. This result is accomplished by using the signal pairs coherence as the main feature of the classifier, rather than looking at their separate information, allowing us to reduce the algorithm complexity drastically. 

\begin{table*}[t!]
\centering
 \caption{Comparison between the accuracy of the literature employing the DB2 (Exercise B) dataset for gesture classification. It can be noted that our approach outperforms DL methods, while using a single feature and a shallow classifier (SVM).}
\begin{tabular}{ |c|c|c|c|c| } 
 \hline
  \textbf{Paper} & \textbf{Approach} &  \textbf{Features} & \textbf{N. of gestures}& \textbf{Accuracy} \\
 \hline
  \cite{Eion_RAL2022} & Transfer learning of dilated efficient CapsNet & CNN learned features & 17 &  $78.3\%$ \\
 \hline 
 \cite{Gulati_2021} & LSTM and dilated-CNN & CNN learned features & 17 & $82.0\%$ \\
 \hline
\cite{Sun_2021}  & Temporal dilation in LSTM and DeepNet  & CNN learned features & 17 & $82.4\%$ \\
 \hline
  \cite{Ding_2018} &  Parallel multiple-scale CNN & CNN learned features & 17 & $83.8\%$ \\
 \hline 
 \textbf{This paper} & \textbf{SVM} &  \textbf{MSC} & 17 &  $\textbf{85.1\%}$ \\
 \hline
\end{tabular}
 \label{table:Comparison}
\end{table*}

Our approach, as shown in Table \ref{table:Comparison}, presented an overall accuracy of $85.1\%$. Even when comparing our results with those of the best-performing deep learning method, we can still notice an improvement in the accuracy.  While deep learning and CNNs remove the need for manual feature extraction, they require extensive data and computing hardware. In contrast, we demonstrated that decoding the sEMG signals to get the coherence of the involved muscles reveals significant insights into the CNS's control strategies during hand movements. By leveraging this information, we constructed our features, proving that a simpler method can be just as effective.

\section{CONCLUSION}
\label{sec::Concl}

In this paper, we introduce a novel machine learning classifier that leverages coherence based functional muscle networks to enhance hand gesture perception. By analyzing the orchestration of functional muscles during hand movements, our approach provides a low-dimensional gesture classifier that effectively captures the critical information necessary for accurate gesture execution. The use of the magnitude squared coherence of sEMG signals processed by a shallow SVM classifier demonstrates that this method not only improves classification accuracy but also offers computational efficiency suitable for real-time applications. The results underscore the potential of coherence-based functional muscle networks in decoding neuromuscular coordination, highlighting their importance in developing intuitive and efficient human-machine interfaces. 

Our findings pave the way for advancements in neurorobotics and interactive systems, where real-time, accurate, and efficient hand gesture recognition is crucial. By extracting meaningful features through muscular coherence, the proposed method can be integrated into robotic control strategies, enhancing the precision and adaptability of neural interfaces. Given the multimodal nature of the dataset, future work can refine and improve the classifier by incorporating additional information, further enhancing its robustness and versatility in practical scenarios. Moreover, our results provide deeper insights into the neuromuscular processes involved in gesture execution, supporting the development of advanced and interpretable models for neurorobotic systems and interactive machines. Future work will focus on further refining the model, exploring its applicability to a broader range of gestures and real-world scenarios, and integrating it into practical systems to fully realize its potential.







\bibliographystyle{unsrt}
\bibliography{ArmaniniBIB}
\clearpage

\end{document}